\documentclass[10pt,twocolumn,letterpaper]{article}

\usepackage{cvpr}              

\usepackage{graphicx}
\usepackage{amsmath}
\usepackage{amssymb}
\usepackage{booktabs}
\usepackage{multicol}
\usepackage{colortbl}
\usepackage{color}
\usepackage{xcolor}
\usepackage{rotating}
\usepackage{multirow}
\usepackage{graphicx}

\usepackage[pagebackref,breaklinks,colorlinks]{hyperref}

\usepackage[capitalize]{cleveref}
\crefname{section}{Sec.}{Secs.}
\Crefname{section}{Section}{Sections}
\Crefname{table}{Table}{Tables}
\crefname{table}{Tab.}{Tabs.}

\def\cvprPaperID{12510} 
\def\confName{CVPR}
\def\confYear{2025}

\begin{document}

\title{From Diffusion to Resolution: Leveraging 2D Diffusion Models for 3D Super-Resolution Task}

\author{Bohao Chen\\
\and
Yanchao Zhang\\
\and
Yanan Lv\\
\and
Hua Han\\
\and
Xi Chen\\
\vspace{-3.0em}
\and
{\small School of Advanced Interdisciplinary Studies, University of Chinese Academy of Sciences}\\
{\small Institute of Automation, Chinese Academy of Sciences}\\
{\small School of Artificial Intelligence, University of Chinese Academy of Sciences}\\
{\small Key Laboratory of Brain Cognition and Brain-inspired Intelligence Technology, Institute of Automation, Chinese Academy of Sciences}\\
{\small School of Future Technology, University of Chinese Academy of Sciences}\\
\vspace{-2.0em}
}

\author{Bohao Chen$^{1,2}$ \quad Yanchao Zhang$^{2,3}$ \quad Yanan Lv$^{2,3}$ \quad Hua Han$^{2,3,4,5}$ \quad Xi Chen$^{2}$\\
{\tt \small \{chenbohao2024, zhangyanchao2021, lvyanan2018, hua.han, xi.chen\}@ia.ac.cn}\\
{\small $^1$School of Advanced Interdisciplinary Studies, University of Chinese Academy of Sciences}\\
{\small $^2$Institute of Automation, Chinese Academy of Sciences}\\
{\small $^3$School of Artificial Intelligence, University of Chinese Academy of Sciences}\\
{\small $^4$Key Laboratory of Brain Cognition and Brain-inspired Intelligence Technology, Institute of Automation, Chinese Academy of Sciences}\\
{\small $^5$School of Future Technology, University of Chinese Academy of Sciences}\\
\vspace{-2.0em}
}

\maketitle

\begin{abstract}

Diffusion models have recently emerged as a powerful technique in image generation, especially for image super-resolution tasks. While 2D diffusion models significantly enhance the resolution of individual images, existing diffusion-based methods for 3D volume super-resolution often struggle with structure discontinuities in axial direction and high sampling costs. In this work, we present a novel approach that leverages the 2D diffusion model and lateral continuity within the volume to enhance 3D volume electron microscopy (vEM) super-resolution. We first simulate lateral degradation with slices in the XY plane and train a 2D diffusion model to learn how to restore the degraded slices. The model is then applied slice-by-slice in the lateral direction of low-resolution volume, recovering slices while preserving inherent lateral continuity. Following this, a high-frequency-aware 3D super-resolution network is trained on the recovery lateral slice sequences to learn spatial feature transformation across slices. Finally, the network is applied to infer high-resolution volumes in the axial direction, enabling 3D super-resolution. We validate our approach through comprehensive evaluations, including image similarity assessments, resolution analysis, and performance on downstream tasks. Our results on two publicly available focused ion beam scanning electron microscopy (FIB-SEM) datasets demonstrate the robustness and practical applicability of our framework for 3D volume super-resolution.

\end{abstract}

\section{Introduction}
\label{sec:intro}

\begin{figure}[!t]
    \centering
    \includegraphics[width=1\columnwidth]{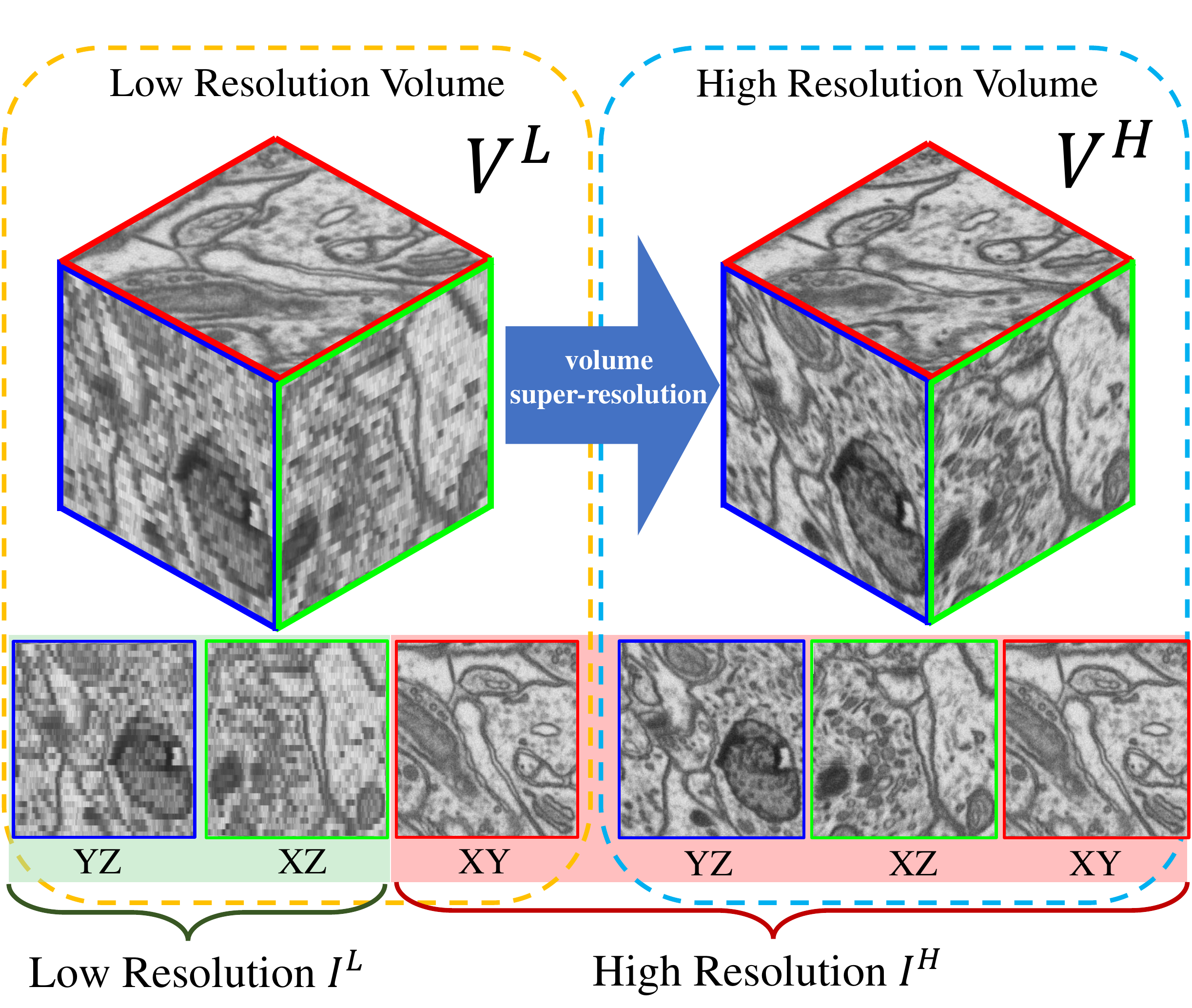}
    \caption{\textbf{Super-Resolution of low-resolution vEM volume.} Biological samples exhibit a consistent data distribution across spatial dimensions. In this work, we propose a training framework named Diffusion to Resolution (D2R) that leverages this intrinsic property to train 3D super-resolution networks without any high-resolution volumes as supervision. Our proposed 3D super-resolution network trained under D2R framework successfully performs 8x super-resolution on low-resolution volumes, achieving performance comparable to that of supervised training.}
\vspace{-1.0em}
\end{figure}

Recent advances in diffusion models have revolutionized image processing, making significant contributions to tasks such as restoration\cite{lugmayr2022repaint, luo2023image}, generation\cite{jain2024video, kim2024arbitrary}, and enhancement\cite{ zhang2020deblurring}. Their powerful generative capabilities are increasingly being applied to improve image quality in fields beyond natural images, such as remote sensing\cite{xiao2023ediffsr, khanna2023diffusionsat}, microscopy\cite{hui2024microdiffusion, lu2024diffusion, moghadam2023morphology}, and MRI/CT imaging\cite{lee2023improving, lyu2022conversion}, where high-resolution data is scarce but critical for downstream tasks. This makes diffusion models particularly promising for volumetric microscopy, where accurate 3D reconstructions require high-resolution volumes, but are hindered by technical and resource limitations \cite{lu2024diffusion, jiang2024super}.

Despite the promising potential of diffusion models for enhancing image quality, their direct application to volumetric electron microscopy (vEM) is fraught with significant challenges. In recent years, vEM has become a pivotal imaging tool in biological research, providing ultrastructural details at nanoscale resolution \cite{knott2013dead}. However, acquiring large-scale isotropic 3D data remains a significant challenge due to the constraints of the imaging method and budget limitations \cite{higginbo2023seven, peddie2022volume}. These constraints underscore an urgent demand for efficient super-resolution methods capable of reconstructing high-quality isotropic volumes from lower-resolution data. 


To address these challenges, we propose Diffusion to Resolution (D2R), a training framework that leverages the 2D diffusion model to train 3D volume super-resolution networks. Inspired by previous works \cite{jiang2024super, deng2020isotropic, he2023isovem, pan2023diffuseir}, D2R is based on the assumption that biological samples exhibit a consistent data distribution across different spatial dimensions. This isotropic assumption enables D2R to leverage 2D diffusion models for learning to recover degraded slices in the XY plane and then apply the learned recovery process to slices in the other two planes. Once trained, the 2D diffusion model generates high-resolution lateral slices on the XZ and YZ planes slice-by-slice, resulting in a lateral high-resolution training volumes for 3D networks. Following this, we train our proposed Deconvolution Gaussian Embedding Attention Network (DGEAN) along lateral directions to effectively capture feature transformation across slices. By leveraging the above workflow, the D2R training framework enables the training of super-resolution networks without isotropic 3D volumes as supervision, while preserving structural continuity across different dimensions. Moreover, D2R can be easily adapted to any volume super-resolution methods that necessitate high-resolution volumes for supervision. Our experiments demonstrate that both DGEAN and other networks trained with the D2R training framework achieve performance comparable to that of networks trained with high-resolution volumes.

In summary, this paper offers the following contributions. 1) We introduce D2R, a novel 2D diffusion-based training framework for the 3D vEM super-resolution task. 2) We propose the DGEAN network, designed to effectively capture high-frequency features within vEM volume data, ensuring seamless interslice continuity and structural coherence. 3) Extensive experiments demonstrated that the DGEAN network integrated with the D2R training framework outperforms all unsupervised vEM super-resolution methods.

\section{Related Work}

\subsection{Volume Super-resolution}
Volume super-resolution (VSR) is a class of techniques used to enhance the axial resolution of a 3D volume, and it has been applied in both electron microscopy (EM) and light microscopy (LM). Classic interpolation methods, such as linear and cubic interpolation, are frequently applied to enhance the axial resolution as they do not introduce artifical biological structures. Recently, deep learning-based interpolation methods like STDIN \cite{wang2022stdin} have notably enhanced volume resolution in the Z-axis (refered as axial direction), yet it is restricted to a fixed $\times2$ super-resolution. Another kind of methods aims to reconstruct lateral (XZ/YZ) images from anisotropic volume, with methods like IsoRecon \cite{deng2020isotropic} using CycleGAN frameworks for restoration, though this requires separate training for each super-resolution factor. Lateral diffusion-based methods, such as DiffuseIR \cite{pan2023diffuseir} and the method proposed in \cite{lee2023reference}, offer arbitrary scale super-resolution reconstruction by filling in missing content in degraded lateral images using a pre-trained diffusion model, although they struggle with spatial continuity and high computational demands. vEMDiffuse \cite{lu2024diffusion} is an axial diffusion-based interpolation method that integrates information from adjacent slices using channel embeddings for different upsample scale factors. However, by restoring only a single 2D image at a time, it encounters prolonged inference times and discontinuities between neighboring slices. Volume-based methods such as 3D SRUNet \cite{heinrich2017deep} achieving isotropic reconstruction of input degraded volume by using high-resolution volume as supervision, but requiring additional training at different scale factors. By performing detail enhancement on bilinear interpolation results of degraded volumes, IsoVEM \cite{he2023isovem} enables volume super-resolution at arbitrary upscale factor. However, this method assumes that the distribution of low-resolution lateral slices is the same as that of high-resolution axial slices, which limits its generalization. Among them, our network DGEAN aligns with vEMDiffuse \cite{lu2024diffusion} in targeting axial reconstruction but diverges by employing a non-generative approach in inference that can efficiently predict missing slices. 

\subsection{Video Frame Interpolation}
Video frame interpolation (VFI) generates intermediate frames between existing input frames, producing smoother motion and higher frame rates. Since Super-SloMo \cite{jiang2018super} utilized optical flow for VFI, many methods have focused on estimating optical flow to synthesize intermediate frames, with \cite{bao2019depth} and \cite{huang2022real} enhancing this approach by incorporating depth maps and ground truth guided flow, respectively. QVI \cite{liu2020enhanced} and IQ-VFI\cite{hu2024iq} improve interpolation performance by fitting nonlinear motions. Due to the complexity and high computational costs involved in estimating optical flow, methods such as CAIN \cite{choi2020channel} and FLAVR \cite{kalluri2023flavr} bypass optical flow estimation and predict frames directly. CAIN uses a channel attention mechanism, while FLAVR employs a 3D UNet architecture with a self-attention layer for end-to-end frame interpolation. Recently, there has been work focusing on the application of diffusion models in VFI tasks\cite{jain2024video}. Compared to traditional interpolation methods, diffusion-based VFI method is able to synthesize natural and high-quality intermediate frames. However, the imaging process in biological scenes suffers from significant noise compared to natural images, and biological structures exhibit strong nonlinear deformations, which strongly hinder VFI-based methods directly used in biological volume super-resolution tasks.

\subsection{Diffusion Models}
Recently, diffusion models have emerged as a leading innovation in generative modeling. Based on modeling a diffusion process and then learning its reverse, the Denoising Diffusion Probabilistic Model (DDPM) \cite{ho2020denoising} is capable of progressively transforming Gaussian noise into structured signals. Subsequent research has focused on refining the control over the output of these models, with key methods such as classifier-guidance \cite{dhariwal2021diffusion} and classifier-free guidance \cite{ho2022classifier}. In image restoration, the core task is to reconstruct high-quality images from corresponding degraded inputs. Common restoration task in natural image field includes image deblurring \cite{zhang2020deblurring}, inpainting\cite{lugmayr2022repaint}, super-resolution\cite{luo2023image, xiao2023ediffsr} and so on. VSR task is similar to lateral slice-by-slice super-resolution tasks, but our approach fundamentally differs from diffusion-based methods. First, due to long sampling time, diffusion models are computationally expensive for VSR. Secondly, vEM slices are noisy, making diffusion-based denoising networks highly sensitive to this noise, especially in latent space super-resolution methods such as \cite{kim2024arbitrary}. Lastly, due to the lack of high-resolution volumes for supervision, performing 2D slice-by-slice inference often introduces discontinuities and artifacts, which strongly degrade performance in downstream tasks. In contrast, our framework utilizes 3D convolutional networks trained on diffusion-enhanced data, eliminating inference time overhead. We incorporate a low-frequency loss for supervision to reduce the impact of noise and leverage the inherent continuity within the volume to enhance spatial consistency in 3D structures.

\section{Proposed Method}

\begin{figure*}[t]
\centering
\includegraphics[width=2\columnwidth]{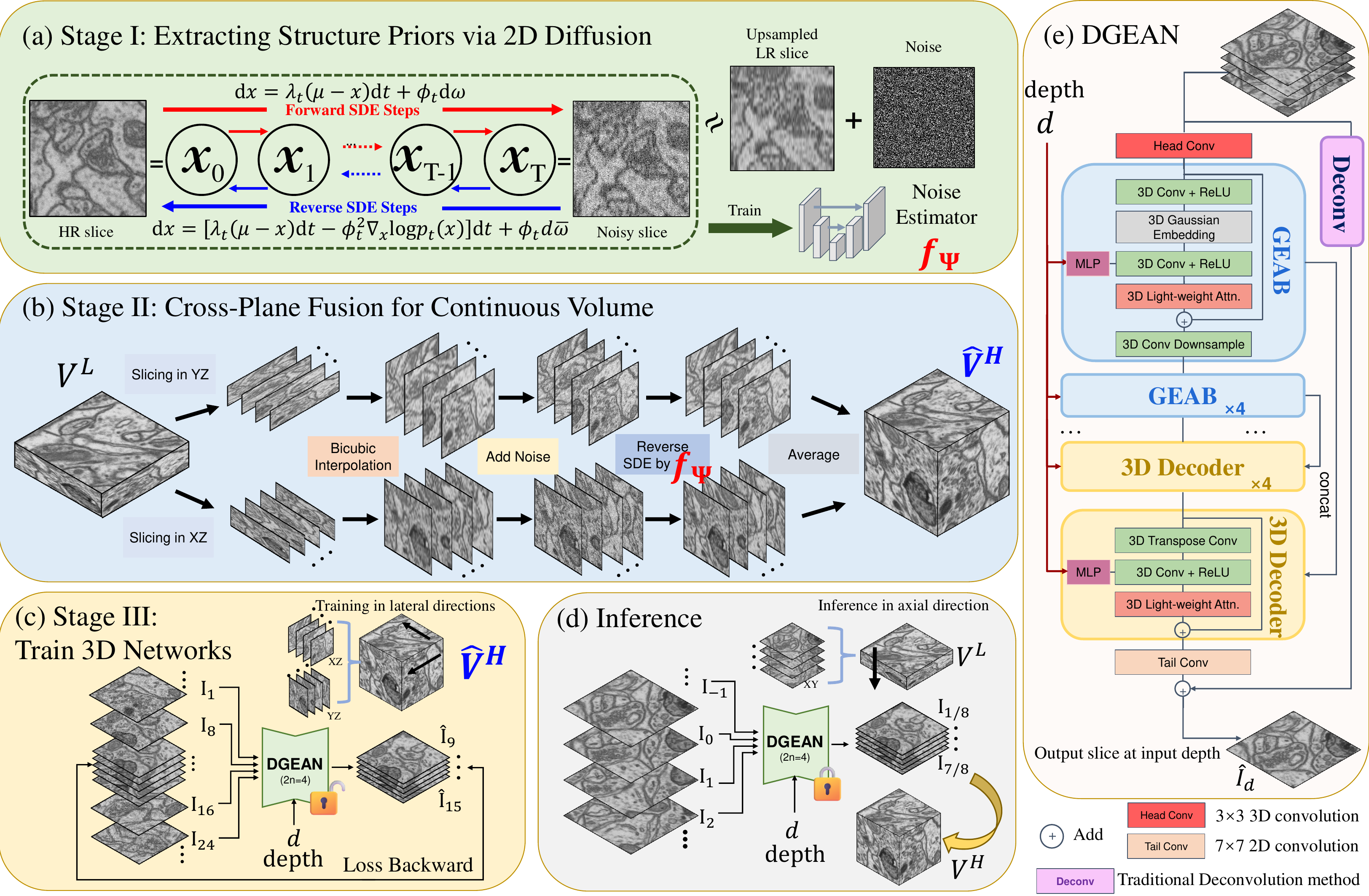} 
\caption{An overview of the proposed D2R training framework and the DGEAN architecture. In Stage I, a 2D diffusion model is trained to restore high-resolution slices from degraded inputs. In Stage II, the trained model recovers the full volume in both XZ and YZ directions. This recovered volume serves as training data for the 3D convolution network in Stage III. After training in Stage III, the 3D convolution network performs inference on the XY plane to recover the high-resolution volume. The DGEAN network, a 3D convolutional network designed for volume super-resolution, shows excellent performance both with high-resolution volumes as supervision and with the D2R training framework, which operates without any high-resolution volume as supervision.}
\label{figure:main_figure}
\vspace{-1.0em}
\end{figure*} 

\subsection{Overview}

 As shown in \cref{figure:main_figure}, our goal is to generate a high-resolution volume $V^{H}$ with increased resolution along the Z-axis by a scale factor $r$. Starting from a low-resolution input volume $V^{L}$, the output $\hat{V}^{H}$ should approximate the ground truth $V^{H}$ closely. Without loss of generality, we set $r = 8$ in this work. We use $I_{XY}$ to denote a slice in the axial direction, where $I_{i;XY} = V_{[i, :, :]}$, and similarly, $I_{j;XZ}$ and $I_{k;YZ}$ represent $V_{[:, j, :]}$ and $V_{[:, :, k]}$, respectively.

 To achieve Z-axis super-resolution without high-resolution volume as supervision, we propose a three-stage 2D diffusion-based training framework named Diffusion to Resolution (D2R). In Stage I, we leverage a 2D diffusion model to learn how to recover the high-resolution slice $I^H_{XY}$ from its low-resolution one $I^{L}_{XY}$ in axial direction. The model is trained on synthesized pairs of degraded data $\hat{I}^{L}_{XY}$ and corresponding high-resolution data $I^{H}_{XY}$. Inspired by previous work \cite{pan2023diffuseir, lee2023reference, lee2024reference, jiang2024super}, in Stage II, we apply the trained diffusion models to recover lateral slices along both the XZ and YZ dimensions. The results are averaged to generate $\hat{V}^{H}$. This volume $\hat{V}^{H}$ is then used as training volume for the 3D network. In stage III, we train a 3D convolution network on the recovery volume $\hat{V}^{H}$ in lateral directions, which allows 3D convolution network to learn stable structure deformation between continuous slices. Finally, the trained 3D model performs super-resolution on the low-resolution volumes along the axial direction, resulting in high-resolution volumes for downstream tasks.
 
\subsection{Extracting Structure Priors via 2D Diffusion}\label{sec:irsde}
The core motivation behind the D2R training framework lies in the assumption that biological structures across different spatial dimensions share an equivalent distribution. This enables training a 2D network in the axial direction and applying it to recover degraded slices in other lateral directions \cite{he2023isovem, pan2023diffuseir, lee2024reference, lee2023reference, jiang2024super}. In Stage I, we focus on recovering the high-resolution slices from the synthesized degraded slices on the XY plane. 

Given a low-resolution 2D image \( I^L \in \mathbb{R}^{H \times W} \), our goal is to predict the corresponding high-resolution slice \( I^H \in \mathbb{R}^{rH \times W} \). Since directly modeling \( p(I^H_{XZ}|I^L_{XZ}) \) and \( p(I^H_{YZ}|I^L_{YZ}) \) becomes challenging due to no high-resolution volumes as ground truth, we approximate the conditional probability of the axial high-resolution slices and the corresponding degraded ones as follows:
\begin{equation}
    p(I^{H}_{XZ} | I^{L}_{XZ}) = p(I^{H}_{YZ} | I^{L}_{YZ})  \approx  p(I^{H}_{XY} | \hat{I}^{L}_{XY}) ,
\end{equation}
where $\hat{I}^{L}_{XY}=\mathrm{downsample}(I^{H}_{XY})$. The downsample process is described in detail in \cref{sec:deg_process}. Here, we employ a light-weight version \cite{xiao2023ediffsr} of IRSDE \cite{luo2023image} to learn above posterior distribution. IRSDE is a diffusion model based on stochastic differential equation (SDE) and outlining the sample generation process using reverse-time SDE. 

The forward diffusion process aims to gradually transform the initial high-resolution slice $I^H = x_0$ to a noisy slice $x_T$ after $T$ time steps, where $x_T \approx \text{upsample}(I^L) + \epsilon, \epsilon \sim \mathcal{N} (0, \delta^2)$. The forward diffusion process is described as follows:
\begin{equation}
    \mathrm{d}x=\lambda_t(\mu-x)\mathrm{d}t + \phi_t \mathrm{d} \omega,
    \label{equ:IRSDE_process}
\end{equation}
where $\mu$ is the state mean, $\omega$ refers to a standard Wiener process, $\lambda_t$ and $\phi_t$ are two time-dependent parameters that control the speed of mean reversion and stochastic volatility, respectively. The $\delta^2$ is set to $\phi^2_t / (2 \lambda_t)$ during training. For a state $x_t$ at $t \in [0, T]$, it can be expressed by \cref{equ:IRSDE_process}:
\begin{equation}
    x_t = \mu + (x_0 - \mu)e^{-\bar{\lambda}_t} + \int_0^t \phi_z e^{-\bar{\lambda}_t} \mathrm{d}\omega(z), 
\end{equation}
where $\bar{\lambda_t}$ is equal to $\int^t_0 \lambda_z \mathrm{d}z$. State $x_t$ follows a Gaussian distribution given by $x_t \sim p_t(x) = \mathcal{N}(x_t | m_t(x), n_t)$, where $m_t(x) = \mu + (x_0 - \mu) e^{-\bar{\lambda}_t}$ and $n_t = \delta^2(1 - e^{ -2 \bar{\lambda}_t} )$ are the mean and variance of this Gaussian distribution, respectively. 

Reverse diffusion aims to recover the high-resolution slice from the terminal noisy state $x_T$. In \cite{luo2023image}, the reverse SDE process is as follows: 
\begin{equation}
    \mathrm{d}x = [\lambda_t(\mu-x)\mathrm{d}t - \phi_t^2\nabla_x \log p_t(x)] \mathrm{d}t + \phi_t\mathrm{d}\bar{\omega}
\end{equation}
where $\bar{\omega}$ denotes a reverse-time Wiener process. $\nabla_x \log p_t (x)$ is the ground truth score during inference stage. In the training phase, the ground truth slice \( x_0 \) is available, enabling the use of more informative conditional scores to enhance model training. Specifically, it is defined as $\nabla_x \log p_t (x|x_0) = -(x_t - m_t(x)) / n_t $.

To reparameterize \( x_t \) as \( x_t = m_t(x) + \sqrt{n_t} \sigma_t \), where \( \sigma_t \sim \mathcal{N}(0, I) \), the ground-truth scores are defined as \( -(\sigma_t / \sqrt{n_t}) \). Given known values for \( m_t(x) \) and \( n_t \), the noise can be estimated using a prediction network \( f_\Psi \). To mitigate training instability in diffusion models, IRSDE applies a maximum likelihood approach instead of directly estimating the noise in \cite{ho2020denoising}. The loss function used in IRSDE is as follows:
\begin{equation}
    \mathcal{L}(\phi) = \sum_{t=0}^T \gamma_t \mathbb{E}\left[\left|\left| x_t - (\mathrm{d}x_t)_{f_\Psi} - x^*_{t-1} \right|\right|\right],
\end{equation}
where \( x^*_{t-1} \) is the ideal state derived from \( x_t \) and given by:
\begin{equation}
    x^*_{t-1} = \hat{\lambda}_te^{-\lambda'_t} (x_t - \mu) \\
    + \hat{\lambda}_te^{-\lambda'_{t-1}} (x_0 - \mu) + \mu,
\end{equation}
where $\hat{\lambda}_t = (1 - e^{-2\bar{\lambda}_{t-1}})/(1 - e^{-2\bar{\lambda}_{t}})$. The proof and more details can be referred in \cite{luo2023image}.

\subsection{Cross-Plane Fusion for Continuous Volume}

Once Stage I training is complete, the trained diffusion model can recover the high-resolution slice under low-resolution observation, represented by the posterior distribution \( p(I^H | I^L) \). Following \cite{jiang2024super}, we slice the volume \( V^{L} \) along the XZ and YZ directions, and separately recover $i$-th low resolution slice $I^{L}_{i;\{XZ/YZ\}}$ to its high resolution slice $\hat{I}^{H}_{i;\{XZ/YZ\}}$ independently. The super-resolution process for a single lateral slice is as follows:
\begin{equation}
    \hat{I}^{H}_{i;\{XZ/YZ\}} = f_{\Psi}(\text{upsample}(I^{L}_{i;\{XZ/YZ\}})+\epsilon),
\end{equation}
where $\text{upsample}(\cdot)$ represent bicubic interpolation. The noise term \( \epsilon \sim \mathcal{N}(0, \delta^2) \) and the trained diffusion model \( f_{\Psi}(\cdot) \) are as described in \cref{sec:irsde}. After slice-wise super-resolution, the slices are concatenate along XZ/YZ direction to recover two high-resolution volumes as follows:
\begin{align}
    \hat{V}^H_{XZ} &= \text{concat} \left( \hat{I}^H_{1;XZ},  \ldots, \hat{I}^H_{n_{XZ};XZ} \right), \\
    \hat{V}^H_{YZ} &= \text{concat} \left( \hat{I}^H_{1;YZ} , \ldots, \hat{I}^H_{n_{YZ};YZ} \right),
\end{align}
where $n_{XZ}$ and $n_{YZ}$ represent slice number in corresponding directions, respectively. Finally, the high-resolution volumes from the XZ and YZ planes are averaged to produce the final high-resolution volume \( \hat{V}^H \) for Stage III training:

\begin{equation}
    \hat{V}^H = \frac{\hat{V}^H_{YZ} + \hat{V}^H_{XZ}}{2}.
\end{equation}

\subsection{3D Convolutional Network for VSR}

In stage III, our goal is to recover the intermediate slice between consecutive input slices. Specifically, in a given sequence of \(2n\) slices \( I_{1:2n} = \{I_1, \ldots, I_n, I_{n+1}, \ldots, I_{2n}\} \), our aim is to generate \((r-1)\) slices between \(I_n\) and \(I_{n+1}\) under the super-resolution factor \(r\). For this purpose, we design a network that takes relative depth parameter \(d\) as input, enabling it to generate the intermediate slices \(\hat{I}_d\) required at the input depth \(d\). Specifically, the task can be formulated as $\hat{I}_d = g_\theta(I_{1:2n}, d)$, where \(g_\theta\) is the network function parameterized by \(\theta\). In this paper, we use $2n=4$ for all experiments.

\subsubsection{DGEAN Architecture Overview}

\cref{figure:main_figure} (e) illustrates the architecture of the Deconvolution Gaussian Embedding Attention Network (DGEAN). By employing 3D convolutions, DGEAN captures the continuous spatial structure transformation across adjacent slices. The backbone of DGEAN is a 18-layer 3D ResNet\cite{tran2018closer}, modified to include five feature encoders. To capture high-frequency details within slices, each encoder is equipped with a randomly initialized Gaussian embedding position module and a relative depth encoding module. A lightweight 3D feature attention layer is incorporated after each position-encoding convolution to enhance details. These modules are combined into the Gaussian Embedding Attention Block (GEAB). Decoders in DGEAN reconstruct the target slice using 3D transposed convolutions and ReLU activations. Skip connections between encoders and decoders ensure accurate multiscale feature fusion. Finally, a \(7 \times 7\) 2D convolution is applied to the 3D spatial feature map of the decoder to generate the output 2D feature map, which is added with traditional deconvolution results \cite{de20173d} as output slice.

\begin{table*}[t]
	\centering
	\footnotesize
	\setlength{\tabcolsep}{3.8pt}
	\begin{tabular}{c|c|c|cccccc}
		\toprule
		\multirow{2}{*}{Dataset} & \multirow{2}{*}{Methods} & \multirow{2}{*}{Supervision} &  
		\multicolumn{6}{c}{Metrics} \\ 
		\cmidrule{4-9} 
		& & & PSNR\textsubscript{XY} $\uparrow$ & PSNR\textsubscript{XZ} $\uparrow$ & PSNR\textsubscript{YZ} $\uparrow$ & SSIM\textsubscript{XY} $\uparrow$ & SSIM\textsubscript{XZ} $\uparrow$ & SSIM\textsubscript{YZ} $\uparrow$ \\ 
		\midrule
		\multirow{11}{*}{FIB25}     
		& SRUNet\cite{heinrich2017deep} & \checkmark  & $27.00_{\pm 1.65}$ & $27.24_{\pm 0.46}$ & $27.24_{\pm 0.46}$ & $0.6945_{\pm 0.08}$ & $\textcolor{blue}{0.7198_{\pm 0.02}}$ & $\textcolor{blue}{0.7073_{\pm 0.02}}$ \\
        
		& vEMDiffuse-i\cite{lu2024diffusion} & \checkmark  & $25.86_{\pm 1.73}$ & $26.15_{\pm 0.61}$ & $26.15_{\pm 0.61}$ & $0.6296_{\pm 0.08}$ & $0.6336_{\pm 0.03}$ & $0.6182_{\pm 0.02}$ \\
            
            & Bicubic & $\times$  & $23.04_{\pm 2.36}$ & $22.69_{\pm 0.46}$ & $22.70_{\pm 0.50}$ & $0.5179_{\pm 0.12}$ & $0.5054_{\pm 0.02}$ & $0.4911_{\pm 0.02}$ \\

		& IsoRecon\cite{deng2020isotropic} & $\times$  & $23.99_{\pm 1.80}$ & $24.07_{\pm 0.46}$ & $24.08_{\pm 0.52}$ & $0.5691_{\pm 0.10}$ & $0.5896_{\pm 0.01}$ & $0.5746_{\pm 0.02}$ \\

            & IsoVEM\cite{he2023isovem} & $\times$  & $24.66_{\pm 1.26}$ & $24.96_{\pm 0.36}$ & $24.96_{\pm 0.36}$ & $0.5959_{\pm 0.09}$ & $0.6339_{\pm 0.03}$ & $0.6162_{\pm 0.03}$ \\

            & IRSDE\cite{luo2023image} & $\times$ & $25.53_{\pm 1.52}$ & $25.87_{\pm 0.37}$ & $25.87_{\pm 0.37}$ & $0.5709_{\pm 0.09}$ & $0.6096_{\pm 0.03}$ & $0.5940_{\pm 0.03}$ \\

            & vEMDiffuse-a\cite{lu2024diffusion} & $\times$ & $23.33_{\pm 1.55}$ & $23.67_{\pm 0.46}$ & $23.68_{\pm 0.54}$ & $0.5104_{\pm 0.10}$ & $0.5257_{\pm 0.05}$ & $0.5036_{\pm 0.05}$ \\

            \cline{2-9}

            & D2R-SRUNet & $\times$ & $27.00_{\pm 1.71}$ & $27.23_{\pm 0.45}$ & $27.23_{\pm 0.45}$ & $0.6899_{\pm 0.08}$ & $0.7149_{\pm 0.02}$ & $0.7018_{\pm 0.02}$ \\

            & Sup-DGEAN (ours) & \checkmark & $\textcolor{red}{27.69_{\pm 1.56}}$ & $\textcolor{red}{28.02_{\pm 0.41}}$ & $\textcolor{red}{28.03_{\pm 0.41}}$ & $\textcolor{blue}{0.7049_{\pm 0.07}}$ & $\textcolor{red}{0.7315_{\pm 0.01}}$ & $\textcolor{red}{0.7182_{\pm 0.02}}$ \\

            & D2R-DGEAN (ours) & $\times$ & $\textcolor{blue}{27.57_{\pm 1.58}}$ & $\textcolor{blue}{27.64_{\pm 0.41}}$ & $\textcolor{blue}{27.64_{\pm 0.41}}$ & $\textcolor{blue}{0.7125_{\pm 0.07}}$ & $0.7148_{\pm 0.02}$ & $0.6994_{\pm 0.02}$ \\

		\midrule
		\multirow{11}{*}{EPFL}     

            & SRUNet\cite{heinrich2017deep} & \checkmark  & $25.79_{\pm 0.80}$ & $26.18_{\pm 0.30}$ & $26.18_{\pm 0.33}$ & $\textcolor{blue}{0.6337_{\pm 0.05}}$ & $\textcolor{red}{0.6834_{\pm 0.02}}$ & $\textcolor{red}{0.6621_{\pm 0.02}}$ \\

            & vEMDiffuse-i\cite{lu2024diffusion} & \checkmark  & $24.88_{\pm 0.78}$ & $25.41_{\pm 0.44}$ & $25.41_{\pm 0.45}$ & $0.5505_{\pm 0.05}$ & $0.5983_{\pm 0.03}$ & $0.5736_{\pm 0.03}$ \\

            & Bicubic & $\times$  & $22.23_{\pm 1.42}$ & $22.30_{\pm 0.37}$ & $22.31_{\pm 0.45}$ & $0.4334_{\pm 0.08}$ & $0.4464_{\pm 0.02}$ & $0.4266_{\pm 0.02}$ \\
            
            & IsoRecon\cite{deng2020isotropic} & $\times$  & $22.93_{\pm 0.80}$ & $23.28_{\pm 0.25}$ & $23.28_{\pm 0.29}$ & $0.4945_{\pm 0.05}$ & $0.5322_{\pm 0.02}$ & $0.5064_{\pm 0.03}$ \\

            & IsoVEM\cite{he2023isovem} & $\times$  & $23.43_{\pm 0.51}$ & $23.89_{\pm 0.25}$ & $23.90_{\pm 0.31}$ & $0.4972_{\pm 0.04}$ & $0.5580_{\pm 0.02}$ & $0.5294_{\pm 0.02}$ \\

            & IRSDE\cite{luo2023image} & $\times$  & $24.38_{\pm 0.81}$ & $24.92_{\pm 0.48}$ & $24.92_{\pm 0.49}$ & $0.5037_{\pm 0.06}$ & $0.5651_{\pm 0.03}$ & $0.5399_{\pm 0.03}$ \\

            & vEMDiffuse-a\cite{lu2024diffusion} & $\times$  & $23.26_{\pm 1.08}$ & $23.74_{\pm 0.49}$ & $23.74_{\pm 0.52}$ & $0.4972_{\pm 0.07}$ & $0.5304_{\pm 0.04}$ & $0.5063_{\pm 0.04}$ \\

            \cline{2-9}

            & D2R-SRUNet & $\times$  & $25.54_{\pm 0.79}$ & $25.99_{\pm 0.31}$ & $25.99_{\pm 0.31}$ & $0.6190_{\pm 0.05}$ & $\textcolor{blue}{0.6689_{\pm 0.02}}$ & $\textcolor{blue}{0.6473_{\pm 0.03}}$ \\

            & Sup-DGEAN (ours) & \checkmark & $\textcolor{red}{26.40_{\pm 0.83}}$ & $\textcolor{red}{26.49_{\pm 0.21}}$ & $\textcolor{red}{26.49_{\pm 0.21}}$ & $\textcolor{red}{0.6435_{\pm 0.04}}$ & $0.6504_{\pm 0.02}$ & $0.6232_{\pm 0.03}$ \\

		& D2R-DGEAN (ours) & $\times$ & $\textcolor{blue}{26.27_{\pm 0.84}}$ & $\textcolor{blue}{26.37_{\pm 0.23}}$ & $\textcolor{blue}{26.37_{\pm 0.23}}$ & $0.6324_{\pm 0.05}$ & $0.6367_{\pm 0.03}$ & $0.6104_{\pm 0.03}$ \\
        
		\bottomrule
	\end{tabular}
    \caption{Similarity metrics comparison on FIB-25 and EPFL datasets. The best and the second-best results are highlighted in \textcolor{red}{red} and \textcolor{blue}{blue}. The supervision column indicates whether the method is supervised (\checkmark) or unsupervised ($\times$) by high-resolution volumes. }
    \label{tab:main_table}
\end{table*}

\subsubsection{Loss functions}\label{sec:loss_functions}
The loss function used to train DGEAN is defined as:
\begin{equation}
    L_\mathrm{total} = L_1 + L_\mathrm{SSIM} + \lambda_\mathrm{FFL} L_\mathrm{FFL} + \lambda_\mathrm{cont} L_\mathrm{cont}
\end{equation}
where $\lambda_\mathrm{FFL}$ and $\lambda_\mathrm{cont}$ are the weights for focal frequency loss \(L_\mathrm{FFL}\) \cite{jiang2021focal} and continuous loss \(L_\mathrm{cont}\), respectively. Low-frequency information is supervised by \(L_1\) and \(L_\mathrm{SSIM}\), and high-frequency details are supervised by $L_{\mathrm{FFL}}$. To promote smoother transitions between slices, we incorporate continuity loss \cite{he2024aid}.
The total continuity loss \(L_\mathrm{cont}\) is the sum of the consistency loss and the smoothness loss, as $L_{\mathrm{cont}} = L_{\mathrm{consist}} + L_{\mathrm{smooth}}$. Using Learned Perceptual Image Patch Similarity (LPIPS) \cite{zhang2018unreasonable, lpipsmodule} as the perceptual feature extractor \( P \), the consistency loss \( L_{\mathrm{consist}} \) is defined with scale factor \( r \) as:
\begin{equation}
    L_{\mathrm{consist}}(I_{1:r}; P) = \frac{1}{r-1} \sum_{i=1}^{r-1} P(I_i, I_{i+1}).
\end{equation}

And the smoothness loss $L_{\mathrm{smooth}}$ encourages a continuous and smooth transition between slices. We employ the Gini coefficient $G(X)$ to quantify the discontinuities between slices, which is given by:
\begin{equation}
    G(X) = \frac{\sum_{i=1}^{r}\sum_{j=1}^r|x_i-x_j|}{2r^2\bar{x}},
\end{equation}
where a higher value indicates a more pronounced changes within sequence. The smoothness loss is defined as:
\begin{equation}
    L_{\mathrm{smooth}}(I_{1:r};P) = 1-G(\bigcup^{r-1}_{i=1}P(I_i, I_{i+1})).
\end{equation}

The weights $\lambda_\mathrm{FFL}$ and $\lambda_\mathrm{cont}$ are empirically
set to \( 10^2 \) and \( 0.1 \), respectively.

\section{Experiments}

\subsection{Degredataion Process}\label{sec:deg_process}
In real-world scenario, a vEM slice is modeled as the sum of a noise-free term \( x_v \) and a Poisson-Gaussian noise term \( \epsilon_{PG} \), where \(\epsilon_{PG} = \alpha \epsilon_P + \epsilon_G\) with \(\epsilon_P \sim \text{Poisson}(x / \alpha)\) and \(\epsilon_G \sim \mathcal{N}(0, \sigma^2)\) \cite{wang2004image}. This noise can be approximated as signal-dependent Gaussian noise \cite{foi2008practical}, leading to a simplified expression as :
\begin{equation}
    y_v = x_v + \epsilon_G', \quad \epsilon_G' \sim \mathcal{N}(0, \alpha x + \sigma^2).
\end{equation}

Previous methods often simulate anisotropic vEM slice $\hat{y}$ by mean-downsampling along the Z-axis, which reduces noise variance by \(1/r\):
\begin{equation}
    \hat{y} = \frac{1}{r} \sum_{i=1}^r y_i = \hat{x} + \hat{\epsilon},\  \hat{x} = \frac{1}{r} \sum_{i=1}^r x_i, \  \hat{\epsilon} \sim \mathcal{N} \left( 0, \frac{1}{r} (\alpha \hat{x} + \sigma^2) \right).
\end{equation}

In our experiments, we apply a downsampling method used in \cite{lu2024diffusion} that discards intermediate slices to create an anisotropic volume. With scale factor $r$, every \(r\)-th slice in axial direction of high-resolution $V^H$ is retained in the low-resolution volume \(V^{L}\), and others are discarded. This process reduces the axial voxel size by a factor of \(r\) compared to the original volume.

\begin{figure*}[ht!]
    \vspace{-2.0em}
    \centering
    \begin{minipage}{0.13\textwidth} 
        \centering
        \includegraphics[height=9.8cm]{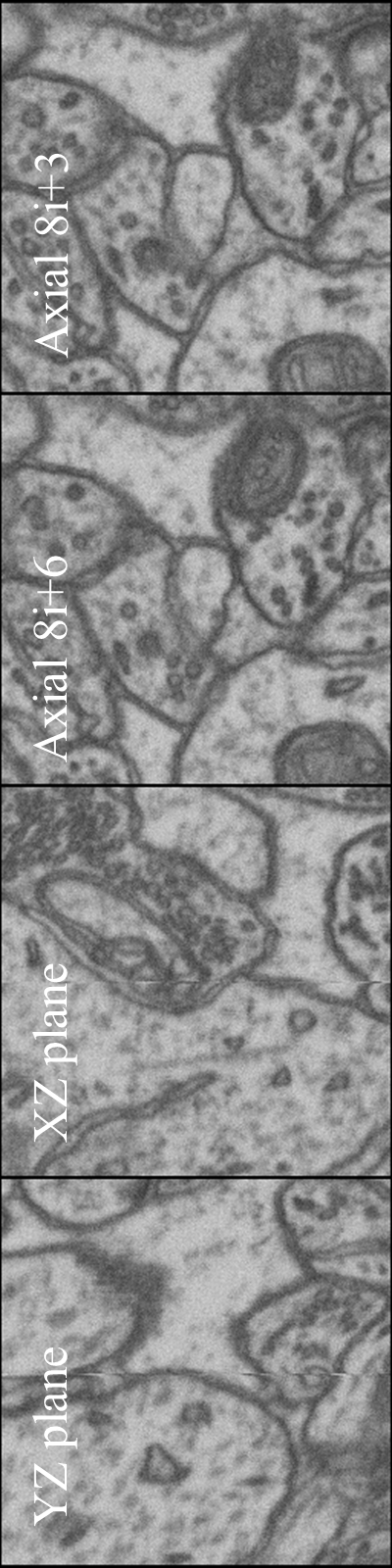}
        \caption*{Ground Truth} 
    \end{minipage}%
    \hspace{0.1cm}
    \begin{minipage}{0.13\textwidth}
        \centering
        \includegraphics[height=9.8cm]{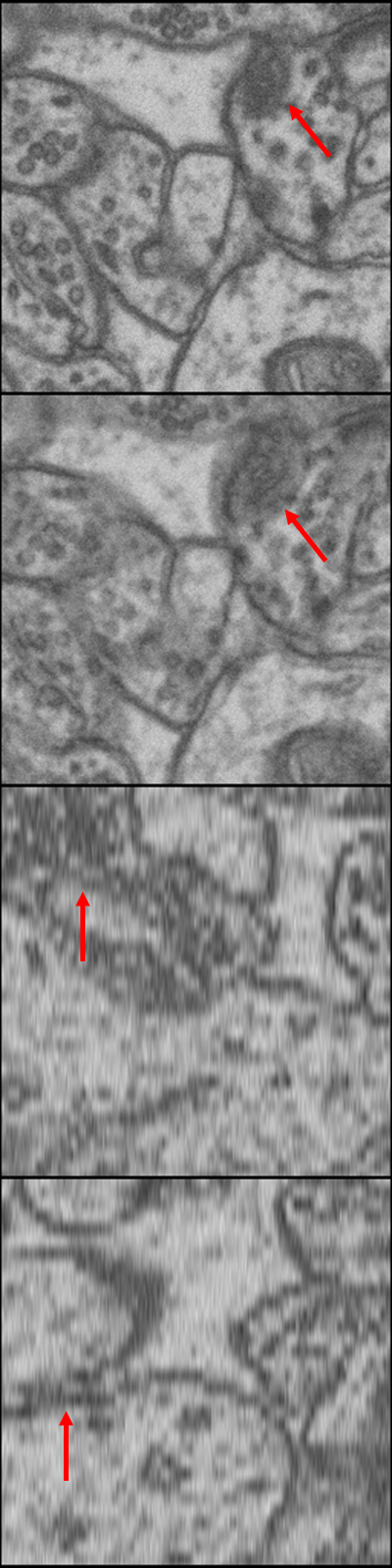}
        \caption*{Bicubic}
    \end{minipage}%
    \hspace{0.1cm}
    \begin{minipage}{0.13\textwidth}
        \centering
        \includegraphics[height=9.8cm]{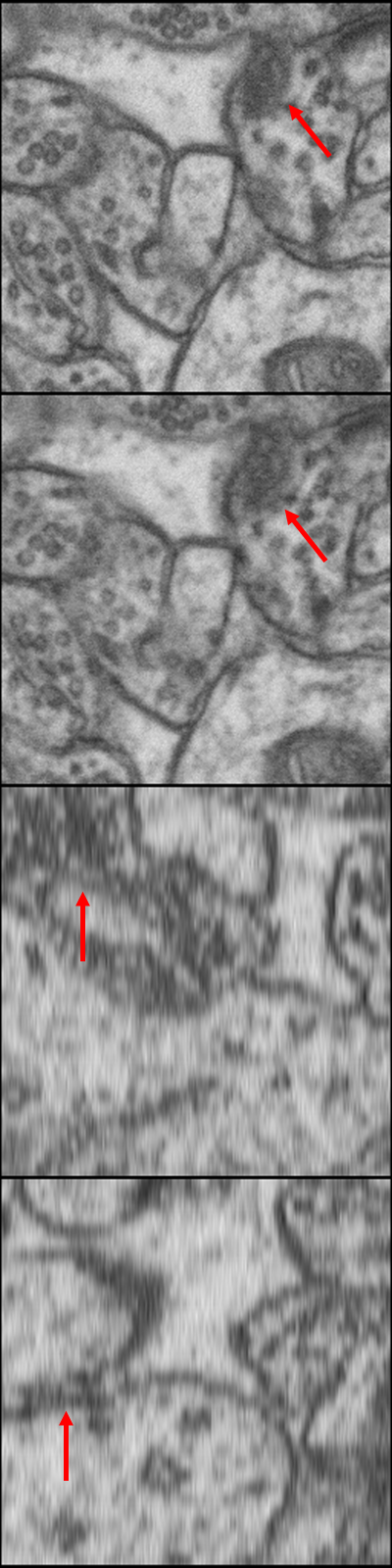}
        \caption*{IsoRecon\cite{deng2020isotropic}}
    \end{minipage}%
    \hspace{0.1cm}
    \begin{minipage}{0.13\textwidth}
        \centering
        \includegraphics[height=9.8cm]{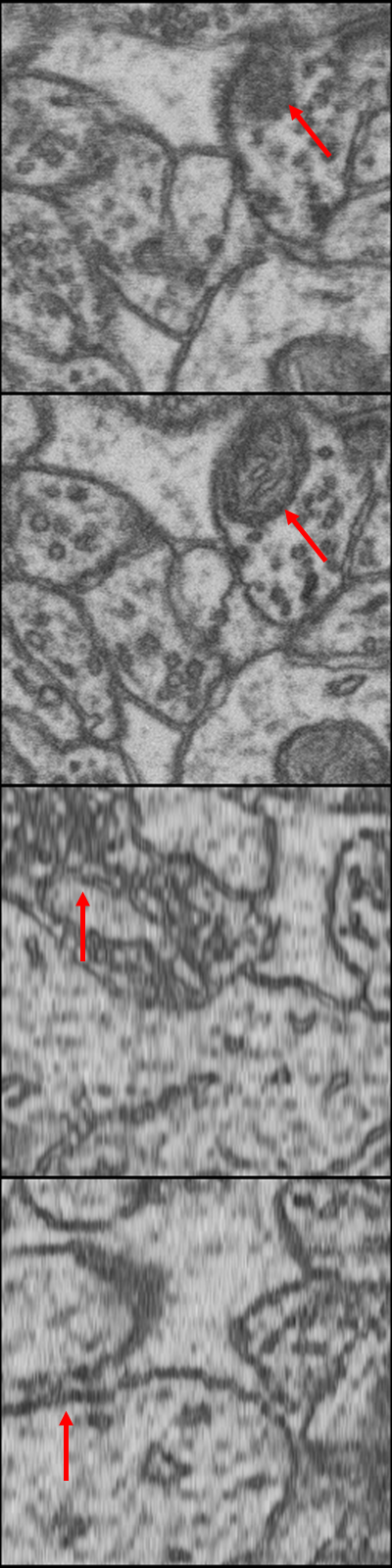}
        \caption*{IsoVEM\cite{he2023isovem}}
    \end{minipage}%
    \hspace{0.1cm}
    \begin{minipage}{0.13\textwidth}
        \centering
        \includegraphics[height=9.8cm]{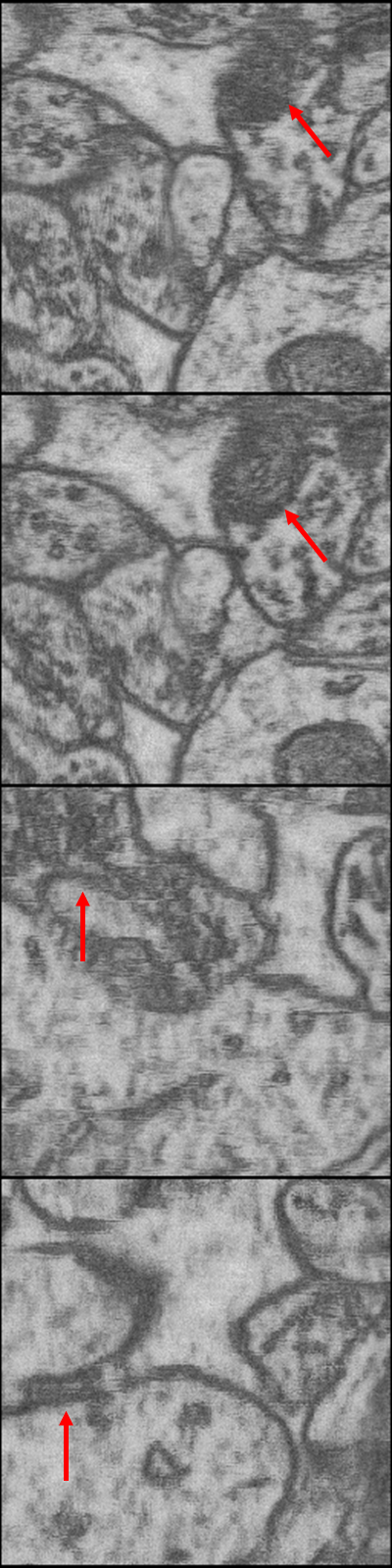}
        \caption*{{\footnotesize vEMDiffuse-a}\cite{lu2024diffusion}}
    \end{minipage}%
    \hspace{0.1cm}
    \begin{minipage}{0.13\textwidth}
        \centering
        \includegraphics[height=9.8cm]{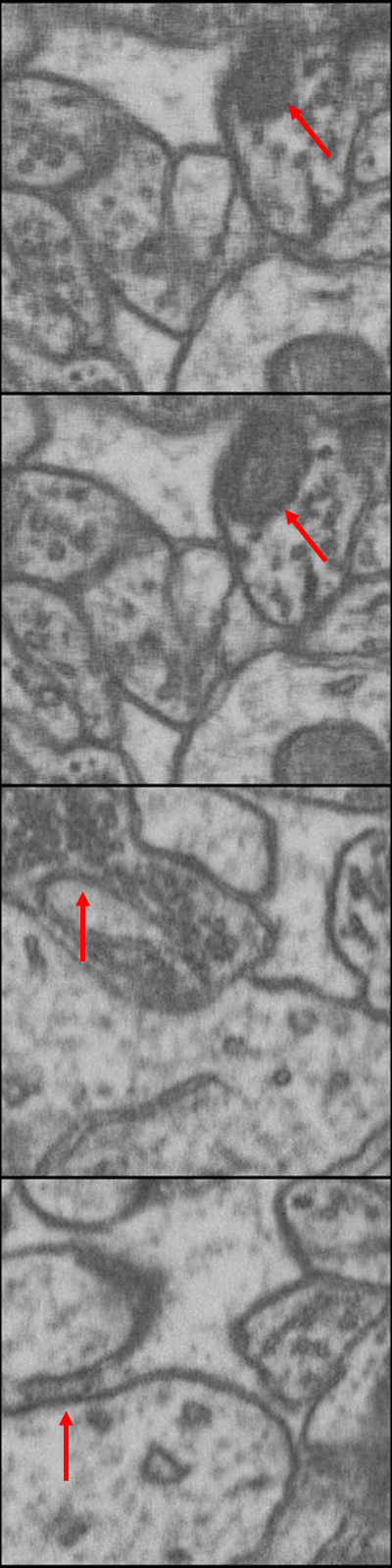}
        \caption*{IRSDE\cite{luo2023image}}
    \end{minipage}%
    \hspace{0.1cm} 
    \begin{minipage}{0.13\textwidth}
        \centering
        \includegraphics[height=9.8cm]{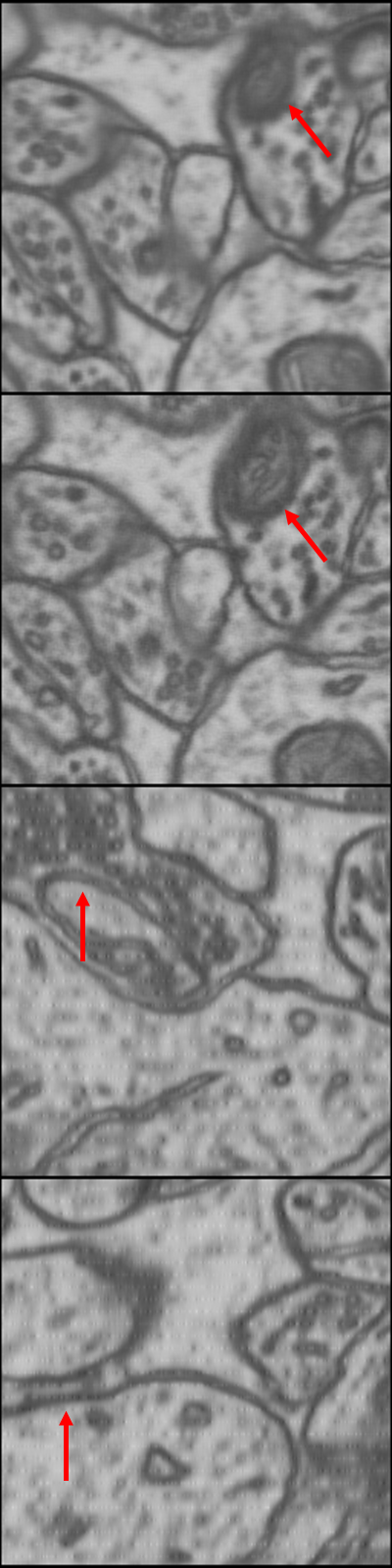}
        \caption*{D2R-DGEAN}
    \end{minipage}
    \caption{Comparison of axial and lateral slices of recovery EPFL volumes using different unsupervised VSR methods.}
    \vspace{-1.0em}
    \label{fig:methods_comparison}
\end{figure*}

\subsection{Datasets}

Since focused ion beam scanning electron microscopy (FIB-SEM) can generate isotropic volumes for nanoscale structures, in our experiments, we evaluated methods on two public FIB-SEM datasets. \textbf{(1) FIB-25}\cite{takemura2015synaptic}: This dataset captures detailed synaptic circuits in the medulla columns of the drosophila visual system, with an isotropic resolution of \(10 \mathrm{nm}\). Five subvolumes were randomly selected for training (80\%) and validation (20\%), each approximately \(2000 \times 2000 \times 512\) voxels. The test set comprises 90 randomly selected subvolumes of \(512\times 512\times512\) voxels. \textbf{(2) EPFL}\cite{epfldataset}: This dataset provided a detailed 3D volume of the CA1 hippocampus region of the mouse brain at an isotropic resolution of \(5 \mathrm{nm}\). The volume was divided into a training set (80\%), a validation set (10\%) and a test set (10\%), with the test set containing 300 subvolumes of \(300\times300\times160\) voxels. Additionally, we selected two separate non-overlapping volume used for resolution estimation and downstream tasks on both FIB-25 and EPFL datasets. With scale factor $r=8$ in our experiments, the axial resolutions of the downsampled FIB-25 and EPFL datasets are 80 nm and 40 nm, respectively.

\subsection{Training Details}
DGEAN is trained using the ADAM optimizer \cite{kingma2014adam} with parameters \(\beta_1=0.9\) and \(\beta_2=0.99\). The initial learning rate is set to \(2\times10^{-4}\) and halved whenever training plateaus. We trained for 60 epochs on each dataset, selecting the model with the best PSNR performance on validation set for further experiments. The training process typically took about 20 hours. The methods with the D2R training framework are trained using their default parameters and training strategies. All experiments were conducted with PyTorch 2.4 on a Linux server with an Intel Xeon Gold 6142 CPU, 512GB RAM, and a Nvidia V100 GPU.

\begin{table}
\centering
\footnotesize
\begin{tabular}{c|c|c}
\toprule
\multirow{2}{*}{Method} & \multicolumn{2}{c}{Estimated Resolution ($\downarrow$)}     \\ 
\cmidrule{2-3}
                        & FIB-25  & EPFL     \\ 
\midrule
SRUNet\cite{heinrich2017deep} & 45.89 nm & 23.79 nm \\
vEMDiffuse-i\cite{lu2024diffusion} & 59.76 nm & 27.05 nm \\
Bicubic                 & 56.48 nm & 31.72 nm \\
IsoRecon\cite{deng2020isotropic} & 58.40 nm & 41.45 nm \\
IsoVEM\cite{he2023isovem} & 54.68 nm & 27.63 nm \\
IRSDE\cite{luo2023image}            & 61.92 nm & 27.63 nm \\
vEMDiffuse-a\cite{lu2024diffusion} & 65.06 nm & 30.59 nm \\
\midrule
D2R-SRUNet & \color{red}{44.31} nm & \color{blue}{22.34} nm \\ 
Sup-DGEAN (ours) & \color{blue}{44.69} nm & \color{red}{21.96} nm \\
D2R-DGEAN (ours) & 45.48 nm & 22.54 nm \\
\bottomrule
\end{tabular}
\caption{Estimated resolution of recovery volumes across datasets. The axial resolutions of degraded volumes are 80nm and 40nm on FIB-25 and EPFL datasets, respectively. The best and second-best results are highlighted in \textcolor{red}{red} and \textcolor{blue}{blue}.}
\vspace{-2.0em}
\label{tab:resolution}
\end{table}

\begin{table}[t!]
\footnotesize
\centering
\begin{tabular}{c|c|c|c|c}
\toprule
\multirow{3}{*}{Method} & \multicolumn{4}{c}{Dataset}                                           \\
\cmidrule{2-5} 
                        & \multicolumn{2}{c}{FIB-25}        & \multicolumn{2}{c}{EPFL}          \\
\cmidrule{2-5}
                        & IoU ($\uparrow$)& Dice ($\uparrow$)& IoU ($\uparrow$)& Dice ($\uparrow$) \\
\midrule
SRUNet\cite{heinrich2017deep}              & 0.6153          & 0.7618          & 0.6968          & 0.8213          \\
vEMDiffuse-i\cite{lu2024diffusion}            &  0.6472  &  0.7858   & \color{red}{0.7533}    &  \color{red}{0.8593}    \\
Bicubic                 & 0.5383          & 0.6998          & 0.5337          & 0.6960          \\
IsoRecon\cite{deng2020isotropic}                & 0.5970          & 0.7476          & 0.4528          & 0.6234          \\
IsoVEM\cite{he2023isovem}                  & 0.5482          & 0.7081          & 0.6541          & 0.7909          \\
IRSDE\cite{luo2023image}           &   0.6142    & 0.7610    & 0.7092    & 0.8299    \\
vEMDiffuse-a\cite{lu2024diffusion}           &   0.5818    & 0.7356    & 0.6663   &  0.7997    \\
\midrule
Sup-DGEAN (ours) & \color{red}{0.6779} & \color{red}{0.8080} & 0.7436 & 0.8530 \\
D2R-SRUNet & \color{blue}{0.6629} & \color{blue}{0.7973} & 0.6846 & 0.8128 \\
D2R-DGEAN (ours) & 0.6546 & 0.7912 & \color{blue}{0.7507} & \color{blue}{0.8576} \\
\bottomrule
\end{tabular}
\caption{Comparison of membrane segmentation accuracy on reconstructed volumes on both datasets. The best and second-best results are highlighted in \textcolor{red}{red} and \textcolor{blue}{blue}. }
\vspace{-2.0em}
\label{tab:mem_seg}
\end{table}

\subsection{Comparison with State-of-the-art Methods}
DGEAN is compared against bicubic interpolation and following state-of-the-art (SOTA) vEM super-resolution methods: SRUNet \cite{heinrich2017deep}, IsoRecon \cite{deng2020isotropic}, IsoVEM \cite{he2023isovem} and vEMDiffuse \cite{lu2024diffusion}. We use the label 'IRSDE' to represent methods that perform 2D super-resolution slice-by-slice in lateral by IRSDE, which is a specific implementation of lateral diffusion models for vEM super-resolution \cite{pan2023diffuseir, lee2023reference, lee2024reference}. Among these methods, SRUNet and vEMDiffuse-i rely on high-resolution volumes for supervision, whereas other methods are trained without the need for ground truth during the training process. To demonstrate the effectiveness of the D2R training framework, we applied it to SRUNet, resulting in the D2R-SRUNet model. We also compared the results of DGEAN trained with high-resolution volume supervision (Sup-DGEAN) and with the D2R training framework (D2R-DGEAN).  All methods are trained on both training datasets.

\begin{table*}[h!]
\vspace{-1.0em}
\footnotesize
\centering
\begin{tabular}{c|ccccccccc}
\toprule
\multirow{2}{*}{Method}    & \multicolumn{2}{c}{Neuron\_1}        & \multicolumn{2}{c}{Neuron\_2}        & \multicolumn{2}{c}{Neuron\_3}     & \multicolumn{2}{c}{Neuron\_4}        \\
\cmidrule{2-9}
                           & IoU ($\uparrow$) & Dice ($\uparrow$) & IoU ($\uparrow$) & Dice ($\uparrow$) & IoU ($\uparrow$) & Dice ($\uparrow$) & IoU ($\uparrow$) & Dice ($\uparrow$) \\
\midrule
SRUNet\cite{heinrich2017deep}                   & { 0.9242}     & { 0.9606}      & 0.8109           & 0.8956            & 0.8157           & 0.8985                     & \textcolor{red}{0.9432}           & \textcolor{red}{0.9707}            \\

vEMDiffuse-i\cite{lu2024diffusion}               & 0.9148           & 0.9555            & \textcolor{blue}{0.8179}     & \textcolor{blue}{0.8999}      & \textcolor{red}{0.8271}  & \textcolor{red}{0.9054}        & \textcolor{blue}{0.9391}           & \textcolor{blue}{0.9686}            \\

Bicubic                    & 0.3609           & 0.5304            & 0.1815           & 0.3072            & 0.2212           & 0.3622                   & 0.2612           & 0.4142            \\

IsoRecon\cite{deng2020isotropic}                   & 0.2300           & 0.3740            & 0.5078           & 0.6736            & 0.1522           & 0.2642                   & 0.3247           & 0.4902            \\
IsoVEM\cite{he2023isovem}                     & 0.7888           & 0.8819            & 0.3870           & 0.5581            & 0.5947           & 0.7458             & 0.3278           & 0.4938            \\

IRSDE\cite{luo2023image} & 0.9172           & 0.9568            & 0.8007           & 0.8893            & 0.7818           & 0.8775             & 0.8265           & 0.9050            \\

vEMDiffuse-a\cite{lu2024diffusion}                     & 0.8416           & 0.9140            & 0.3812           & 0.5520            & 0.5947           & 0.7458            & 0.7916           & 0.8837            \\

\midrule

D2R-SRUNet                   & 0.9042     & 0.9496      & 0.7970           & 0.8870            & 0.7733           & 0.8721           & 0.8377           & 0.9117            \\

Sup-DGEAN (ours)           & \textcolor{red}{0.9315}     & \textcolor{red}{0.9645}      & 0.8109           & 0.8185            & \textcolor{blue}{0.8195}           & \textcolor{blue}{0.9008}          & 0.9345           & 0.9661            \\

D2R-DGEAN (ours) & \textcolor{blue}{0.9244}           & \textcolor{blue}{0.9607}            & \textcolor{red}{0.8199}           & \textcolor{red}{0.9010}            & 0.7854           & 0.8798            & 0.9086           & 0.9521     \\

\bottomrule
\end{tabular}
\caption{Comparison of reconstruction accuracy on volumes recovered by different methods, with metrics computed by comparing manually labeled neurons to those reconstructed from the recovery volumes. The best and second-best results are highlighted in \textcolor{red}{red} and \textcolor{blue}{blue}.}
\vspace{-1.5em}
\label{tab:reconstruction_acc}
\end{table*}

\noindent\textbf{Similarity Comparison.}\quad
We use DGEAN and other comparative methods to reconstruct downsampled subvolumes from the FIB-25 and EPFL test datasets. The similarity metrics are calculated slice-by-slice in all three directions. As shown in \cref{tab:main_table}, our method outperforms all convolution-based, diffusion-based, and Transformer-based methods with low standard deviations. The results also validate the effectiveness of the D2R training framework. In particular, SRUNet and DGEAN trained with D2R show slightly lower performance than those trained with high-resolution volumes as supervision, but still outperform all other unsupervised methods and vEMDiffuse-i. Comparisons of axial and lateral slices presented in \cref{fig:methods_comparison} show that our method recovers better structures compared to other unsupervised methods on testing datasets. It is worth noting that other unsupervised methods introduce artifacts in axial slices (vEMDiffuse-a, IRSDE) or lateral slices (Bicubic, IsoRecon, IsoVEM, vEMDiffuse-a). The arrows point out that D2R-DGEAN is closer to the true axial and lateral slices compared to all other unsupervision methods.

\noindent\textbf{Resolution Comparison.}\quad
Recently, the Fourier shell correlation (FSC) \cite{nieuwenhuizen2013measuring}, which is used as the gold standard for reliable reconstruction resolution estimation, has been applied to evaluate the resolution of FIB-SEM volumes \cite{he2023isovem,lu2024diffusion}. Here, we use FSC-0.5 to assess the resolution of reconstructed volumes. As shown in \cref{tab:resolution}, Sup-DGEAN improved the axial resolution of the FIB25 dataset from 80 nm to 44.69 nm, and the axial resolution of the EPFL dataset from 40 nm to 21.96 nm. All methods trained with the D2R training framework achieve performance comparable to that of methods trained with high-resolution volume as supervision, while surpassing all unsupervised methods. The resolution metric strongly demonstrates the effectiveness of our method in volumetric super-resolution reconstruction.

\noindent\textbf{Membrane Segmentation Comparison.}\quad
We evaluate the performance of the above methods on the membrane segmentation task. To ensure an unbiased comparison, we used a public pre-trained membrane segmentation model \cite{zhang2024segneuron} that was not fine-tuned on either dataset. This model is applied to both the ground truth volumes and the reconstructed volumes. Membrane segmentation results are then assessed using Intersection over Union (IoU) \cite{zhou2019iou} and Dice coefficient metrics \cite{shamir2019continuous}, comparing against those from the ground truth volumes. The results in \cref{tab:mem_seg} show that DGEAN consistently outperforms the other methods, producing segmentations that are closest to those obtained from the ground truth. The results demonstrates the superior performance of DGEAN in preserving membrane structures and achieving high segmentation accuracy after volumetric super-resolution.

\noindent\textbf{Neuron Reconstruction Comparison.}\quad
For neuron reconstruction, high-quality volumes are critical for building affinity graphs for voxel aggregation. To evaluate the quality of the reconstructed volumes to the aggregation algorithm, we applied an automated neuron segmentation pipeline \cite{beier2017multicut} to reconstruct neurons from the recovery volumes. We compared the reconstructed results with ground truth for four randomly selected neurons labeled by experts on the EPFL test volume. As shown in \cref{tab:reconstruction_acc}, Sup-DGEAN achieves performance comparable to vEMDiffuse-i, while avoiding the long inference time and the need for supervision that are inherent drawbacks of vEMDiffuse-i. Moreover, the DGEAN trained within the D2R training framework (D2R-DGEAN) performs similarly to Sup-DGEAN, demonstrating the excellent data augmentation capability of the D2R training framework. These results highlight the effectiveness of both Sup-DGEAN and D2R-DGEAN in accurately recovering high-quality volumes for reconstruction, demonstrating that the D2R training framework significantly enhances model performance in 3D volume super-resolution, particularly for reconstruction tasks.

\section{Conclusion}

In this work, we propose a novel approach to 3D volume super-resolution that leverages lateral continuity in data by 2D diffusion models. By training a 2D diffusion model in axial direction and inference on laterally degraded images, we generate training volumes which preserve lateral continuity. A high-frequency-aware 3D super-resolution network is then trained to learn spatial transformation across slices, enabling inference in axial direction to enhance axial resolution. Comprehensive evaluations on two FIB-SEM high-resolution volume datasets demonstrate the effectiveness and robustness of our approach, providing a solution to the challenge of training without high-resolution volumes as supervision for 3D vEM super-resolution.

{\small
\bibliographystyle{ieee_fullname}
\bibliography{egbib}
}

\end{document}


\title{Supplementary Materials}

\maketitle

\section{More Details of FSC metrics}
\label{sec:FSC_metrics}

In Section 4.4, we estimate the resolution by Fourier Shell Correlation (FSC) with the FSC-0.5 criterion. Here, we provide additional details on the FSC metric employed in our resolution estimation experiments. For a frequency sequence $\{q\}$, the FSC metric between the ground truth volume $V_{gt}$ and the predicted volume $V_{pred}$ is defined as:  

\begin{equation}
\begin{aligned}
        &FSC(V_{gt}, V_{pred}) =  \\
        &\frac{ \sum_{q_i \in \{q\}}F(V_{gt})_{q_i} \cdot F({V_{pred})}^*_{q_i}}{\sqrt{\sum_{q_{i}\in \{q\}} \left|F(V_{gt})_{q_i}\right|^{2} \cdot \sum_{q_{i}\in \{q\}} \left|F(V_{pred})_{q_i}\right|^{2}}},
\end{aligned}
\end{equation}
where $F(V_{gt})_{q_i}$ and $F(V_{pred})_{q_i}$ are the Fourier transforms of $V_{gt}$ and $V_{pred}$ at frequency $q_i$, respectively. The $F(V_{pred})^*$ is the complex conjugate of the Fourier transforms of $F(V_{pred})$. 

\begin{figure}[!t]
    \centering
    \begin{subfigure}[b]{0.45\textwidth}
        \centering
        \includegraphics[width=\textwidth]{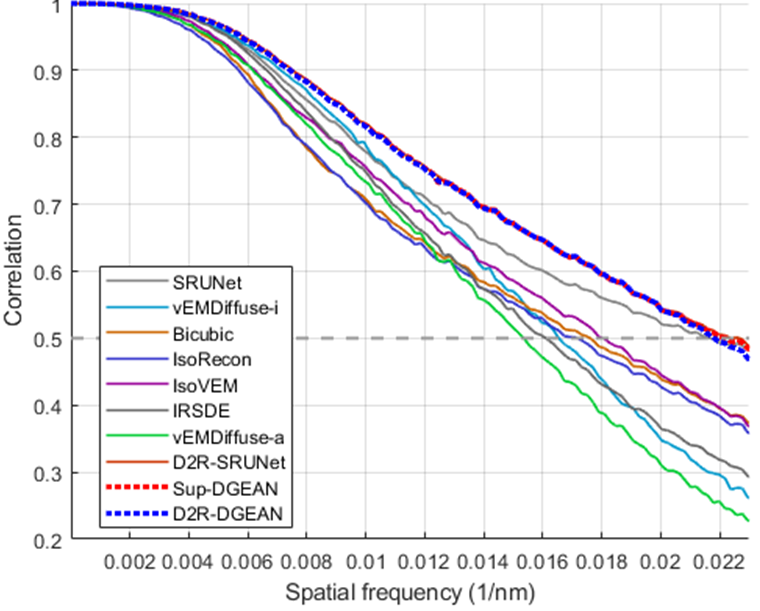}
        \caption{FSC response curves on recovery FIB-25 volumes with ground truth.}
        \label{fig:est_fib25}
    \end{subfigure}
    
    \centering
    \vspace{1em}
    \begin{subfigure}[b]{0.45\textwidth}
        \centering
        \includegraphics[width=\textwidth]{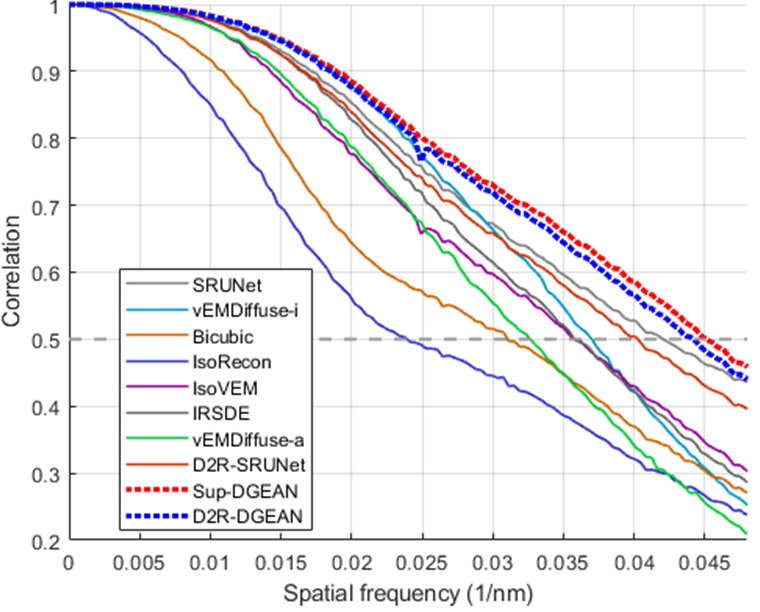}
        \caption{FSC response curves on recovery EPFL volumes with ground truth.}
        \label{fig:est_epfl}
    \end{subfigure}
    \caption{FSC response curves of different methods on FIB-25 and EPFL datasets. The supervised version and unsupervised version of DGEAN are marked in \textcolor{red}{red} dashline and \textcolor{blue}{blue} dashline in both subfigures. An intersection point further to the right along X-axis corresponds to superior resolution performance. Detailed resolution estimations can be found in Tab.\textcolor{red}{2} of the article.}
    \label{fig:est_both_datasets}
\vspace{-1.0em}
\end{figure}

In previous studies \cite{frank2006three, bottcher1997determination}, volume resolution is defined as the spatial frequency at which the FSC curve drops below 0.5, marking the point where the input volumes lose significant agreement. This criterion is widely adopted due to its robustness and ability to quantify the effective resolution of reconstructed 3D data, making it particularly suitable for evaluating super-resolution methods. As mentioned above, in our experiments, we applied the FSC-0.5 criterion to estimate the resolution of recovered volumes produced by different VSR methods, as shown in Table 2 of our acticle. \cref{fig:est_both_datasets} presents the FSC curves for all methods. The selected volumes $V_{gt}$ and $V_{pred}$ are of size $512\times512\times512$ on both datasets.

\section{More Details of Membrane Segmentation}

In Section 4.4, we evaluate the performance of the same membrane segmentation method \cite{zhang2024segneuron} applied to recovery volumes from different volume super-resolution methods by comparing the segmentation results with ground truth volumes using IoU and Dice coefficients. The IoU is defined as:  
\begin{equation}
\text{IoU}(S_{gt}, S_{pred}) = \frac{\left| S_{gt} \cap S_{pred} \right|}{\left| S_{gt} \cup S_{pred} \right|},
\end{equation}
where \( S_{gt} \) and \( S_{pred} \) represent the membrane segmentation results on ground truth and recovery volumes by volume super-resolution methods, respectively. The Dice coefficient is defined as:  
\begin{equation}
\text{Dice}(S_{gt}, S_{pred}) = \frac{2 \left| S_{gt} \cap S_{pred} \right|}{\left| S_{gt} \right| + \left| S_{pred} \right|}.
\end{equation}  

\begin{figure*}[!t]
    \centering
    \begin{subfigure}[b]{0.75\textwidth}
        \centering
        \includegraphics[width=\textwidth]{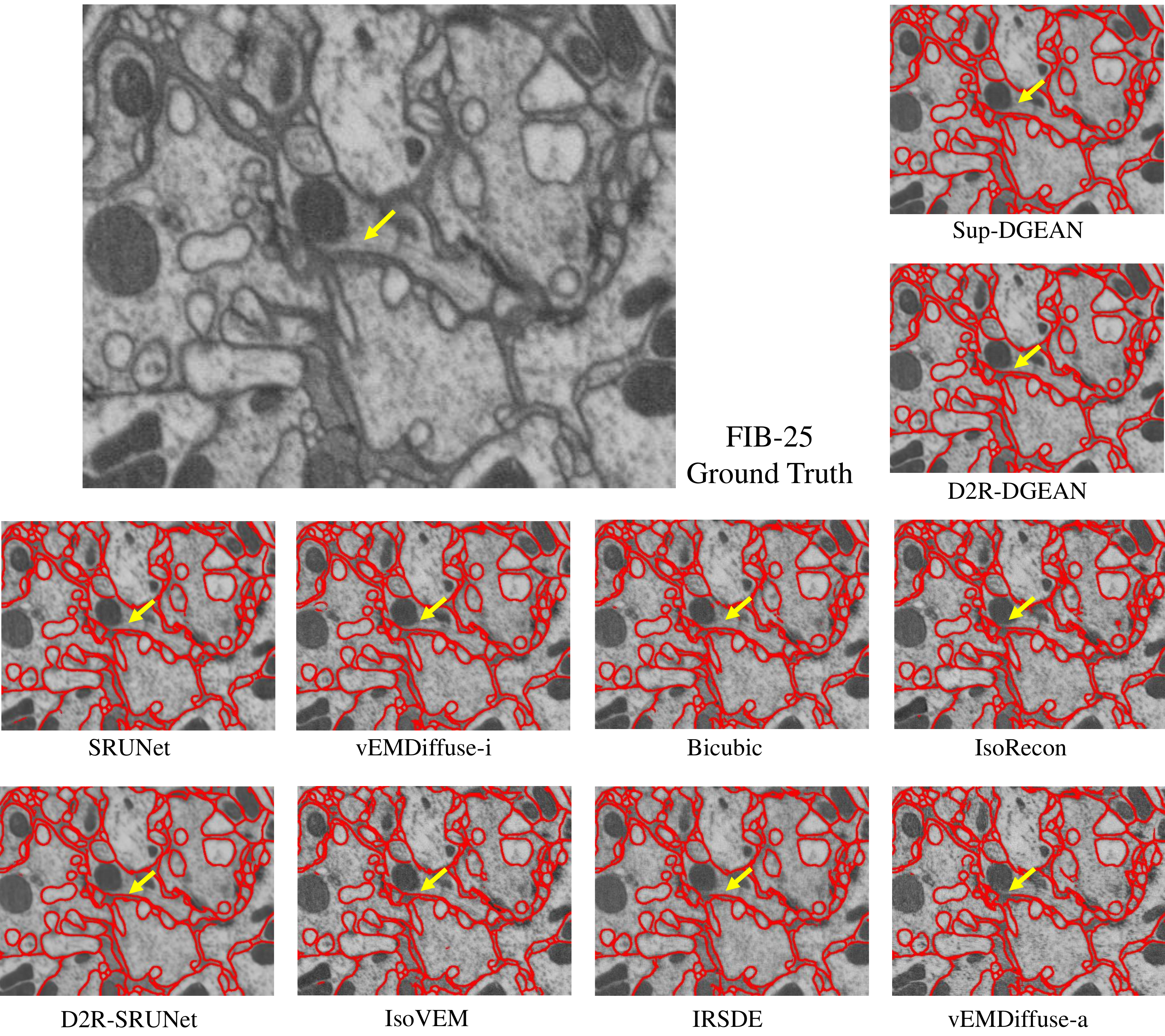}
        \caption{Membrane segmentation on FIB-25 dataset.}
        \label{fig:seg_fib25}
    \end{subfigure}
    
    \centering
    \vspace{1em}
    \begin{subfigure}[b]{0.75\textwidth}
        \centering
        \includegraphics[width=\textwidth]{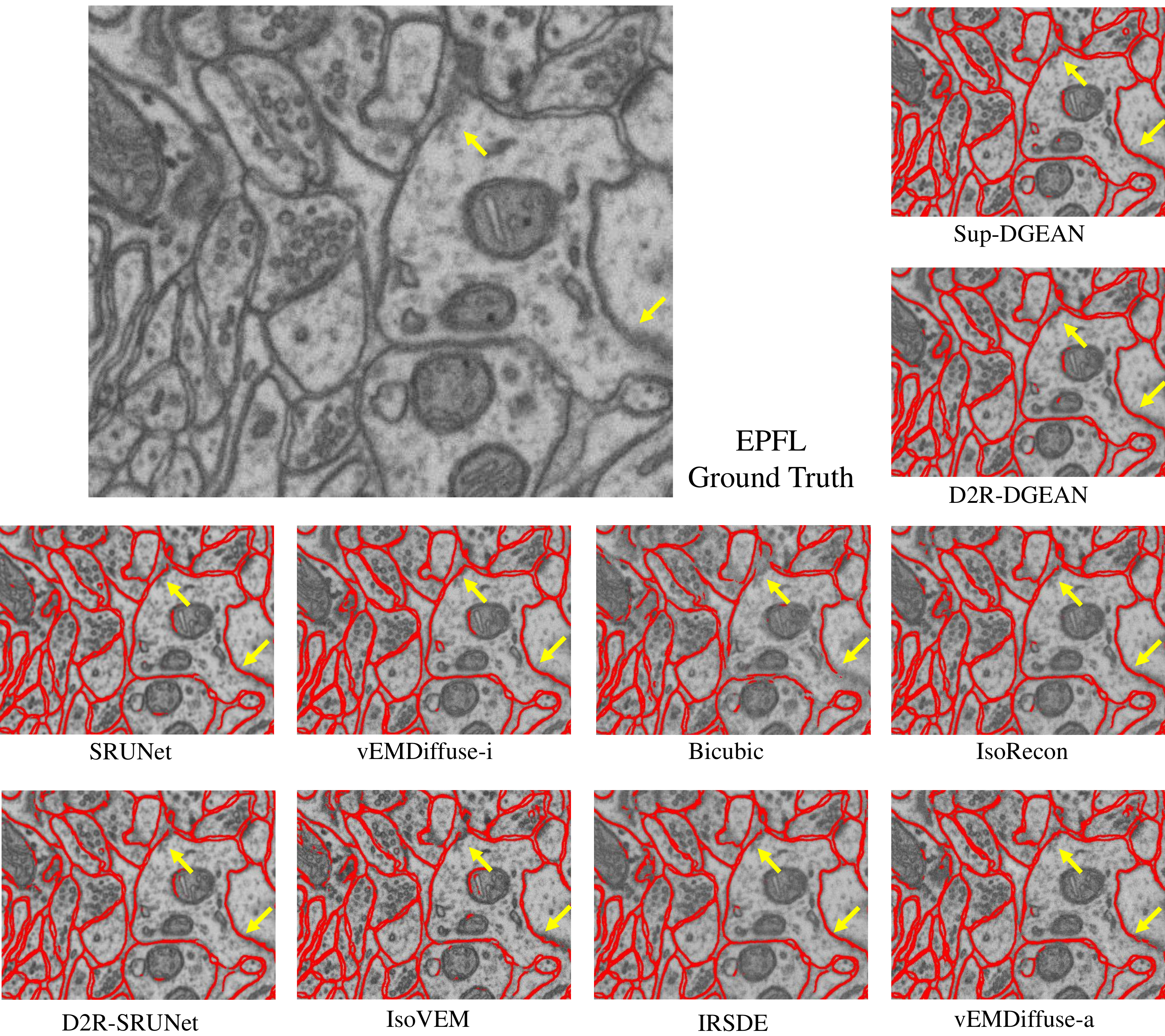}
        \caption{Membrane segmentation on EPFL dataset.}
        \label{fig:seg_epfl}
    \end{subfigure}
    \caption{Membrane segmentation results on different datasets. The quantitative segmentation results can be found in Tab.\textcolor{red}{3} of the article.}
    \label{fig:seg_both_data}
\vspace{-1.0em}
\end{figure*}

In \cref{fig:seg_both_data}, we present the membrane segmentation results for different methods on the FIB-25 and EPFL datasets. Our methods, Sup-DGEAN and D2R-DGEAN, demonstrate superior performance in capturing fine details compared to other competing approaches. Moreover, models trained with the D2R training framework, such as D2R-SRUNet and D2R-DGEAN, achieve segmentation results that are nearly comparable from those trained with high-resolution volume as supervision, such as SRUNet and Sup-DGEAN. This demonstrates the effectiveness and superiority of the proposed D2R training framework in preserving fine details for membrane segmentation task during volume super-resolution.

\section{More Details of Neuron Reconstruction}

\begin{figure*}[htp]
    \centering
    \centering
    \includegraphics[width=\textwidth]{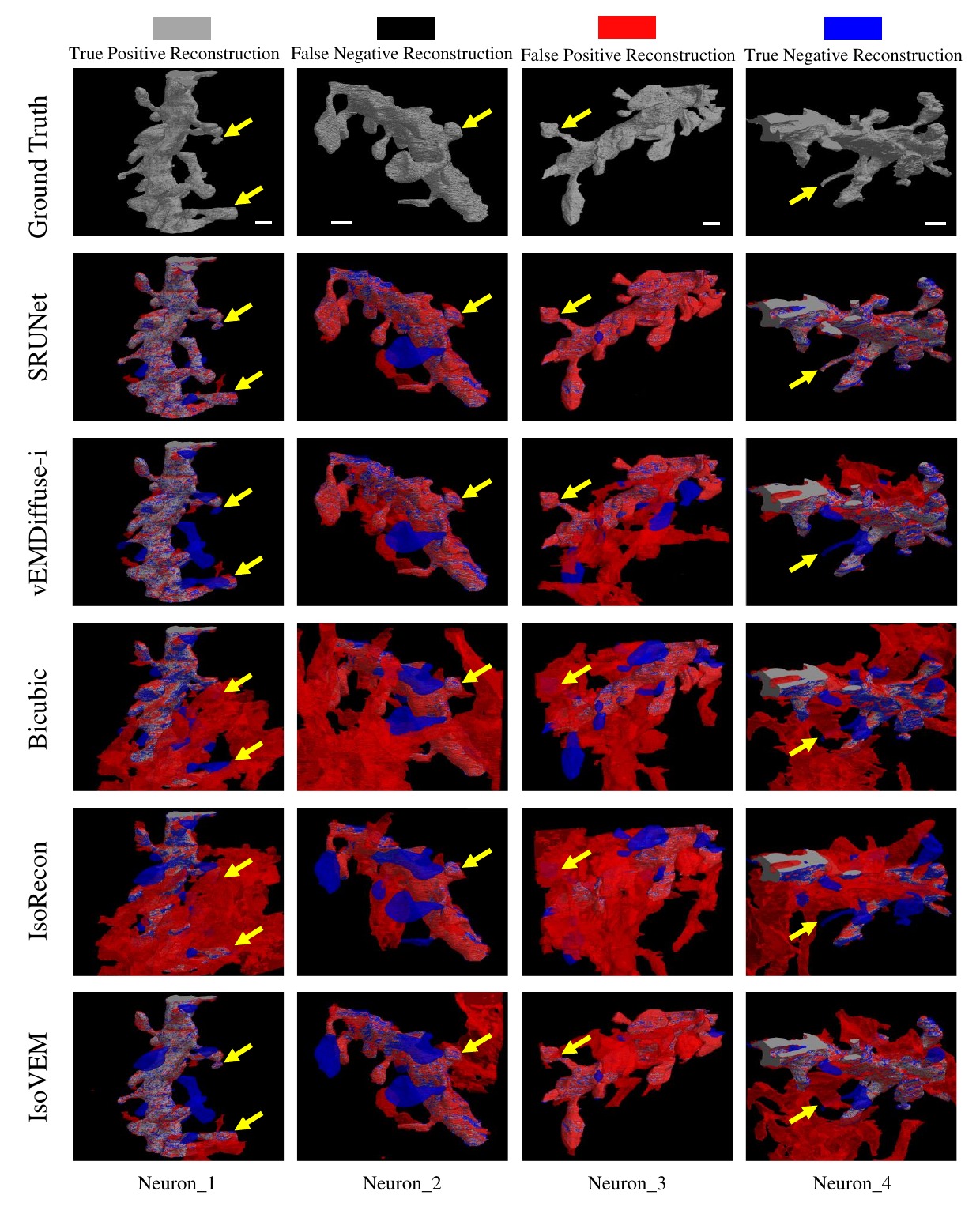}
    \caption{All scale bars in the ground truth row represents 500 nm.}
    \label{fig:large1}
\end{figure*}

\begin{figure*}
    \centering
    \includegraphics[width=\textwidth]{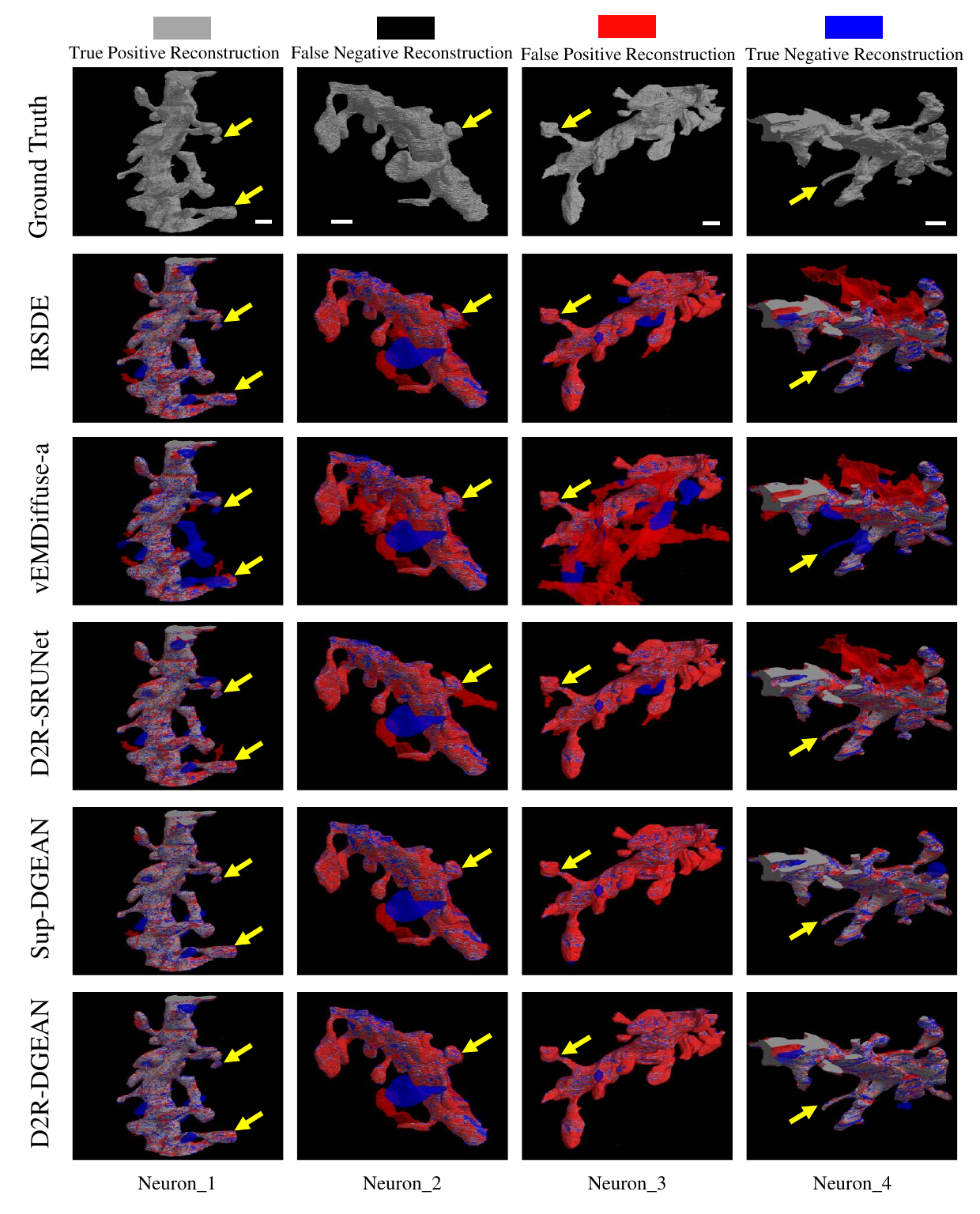}
    \caption{All scale bars in the ground truth row represents 500 nm.}
    \label{fig:large2}
\end{figure*}

In Section 4.4, we evaluate the performance of neuron reconstructions on high-resolution FIB-SEM volume and with recovery volumes, as referenced in Tab.\textcolor{red}{4} of the article. In \cref{fig:large1} and \cref{fig:large2}, we present the neuron reconstruction results of same reconstruction workflow on volumes recovered by different volume super-resolution results. Compared to other methods, results of Sup-DGEAN and D2R-DGEAN achieve higher reconstruction accuracy, with error distributions not correlated with biological structures. Moreover, the recovery volumes enhanced by our methods still allow for relatively continuous tracing of finer structures (marked as yellow arrows), which are often ignored by other methods. Results by models trained with the D2R training framework, achieve reconstruction results that are nearly comparable from those trained with high-resolution volume as supervision, which demonstrates the effectiveness and superiority of the proposed D2R training framework in volume super-resolution task.

\section{Arbitrary-Scale Super-Resolution with DGEAN}

\begin{figure*}
    \centering
    \includegraphics[width=\textwidth]{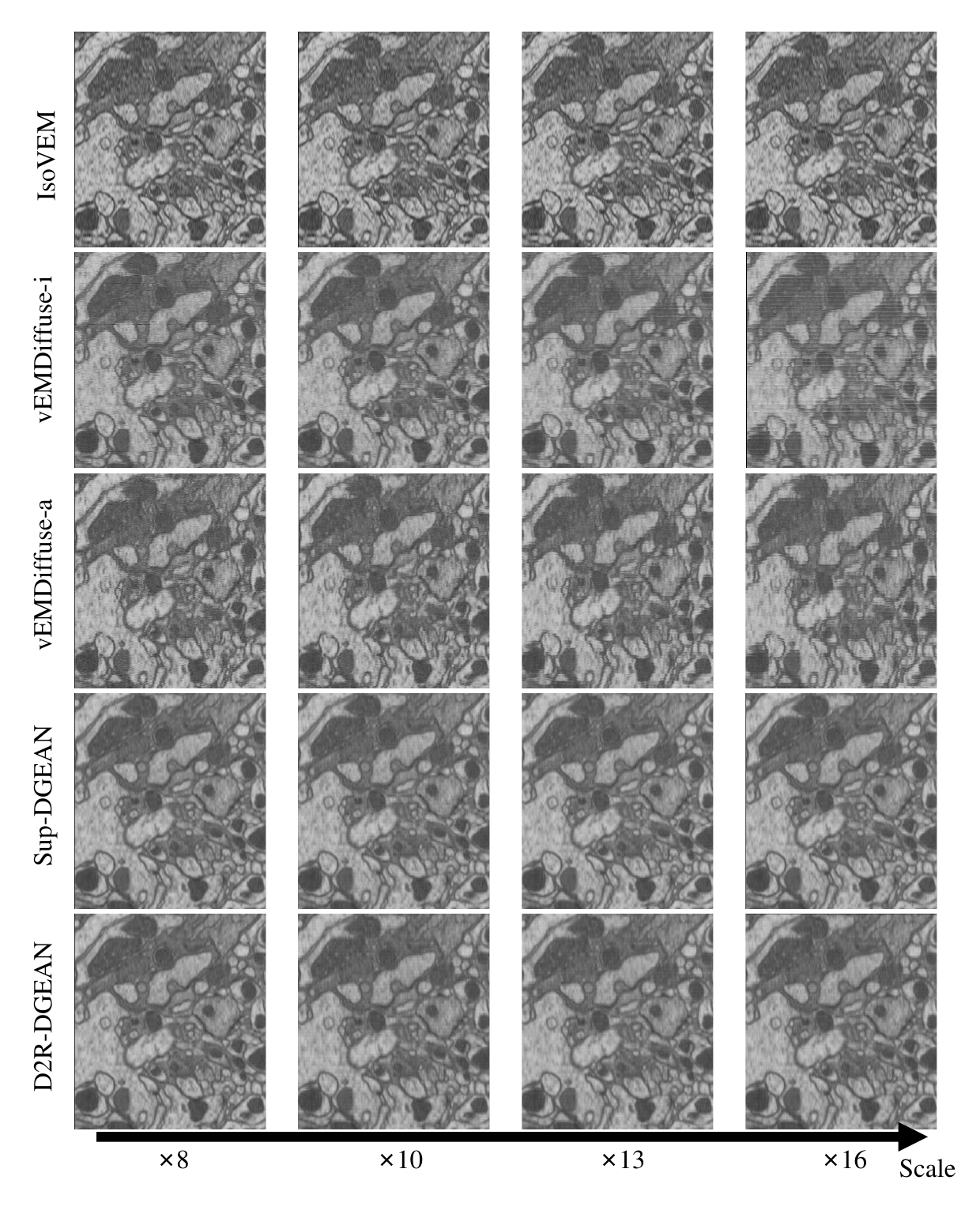}
    \caption{Lateral image of methods at different super-resolution ratios.}
    \label{fig:diff_ratios}
\end{figure*}

By introducing the scaling factor \(d\) as an additional input, DGEAN dynamically adjusts its upsampling and reconstruction processes to generate outputs at any desired resolution. This adaptability eliminates the need for multiple models trained at specific scales, making DGEAN both efficient and versatile for diverse volumetric super-resolution scenarios. Such a feature is particularly advantageous in applications where different magnification levels are required to analyze fine structural details. In \cref{fig:diff_ratios}, we present the super-resolution results of several methods at different upscaling factors. All methods were trained with only \( \times 8 \) super-resolution factor and evaluated without any fine-tuning. The results indicate that the IsoVEM method consistently produces blurry outputs in real-world scenarios with high noise. Meanwhile, both vEMDiffuse methods exhibit noticeable artifacts as the super-resolution factor increases. In contrast, our method demonstrates consistent performance across different super-resolution scales. Moreover, the performance of models trained with high-resolution volume as supervision (Sup-DGEAN) and the D2R training framework (D2R-DGEAN) have same performance, showcasing the effectiveness of both our method and the proposed D2R training framework.

{\small
\bibliographystyle{ieee_fullname}
\bibliography{egbib}
}